%% file: main.tex
\documentclass[letterpaper, 10 pt, conference]{ieeeconf}  

\IEEEoverridecommandlockouts                              

\usepackage{graphicx}
\usepackage{amsmath}
\usepackage{amssymb}
\usepackage{mathtools}
\usepackage{tikz}
\usetikzlibrary{arrows.meta,positioning,fit,shapes.geometric,calc}

\usepackage{kotex}

\usepackage{hyperref}
\usepackage{algorithmic}

\title{\LARGE \bf
Rarity-Gated Context Conditioning for Offline Imitation Learning-Based Maritime Anomaly Detection
}

\author{Yongmin Kim$^{a}$, ByeongHoon Jeon$^{a}$, and Sungil Kim$^{a,\dagger}$%
  \thanks{$^{a}$Department of Industrial Engineering,
    Ulsan National Institute of Science and Technology (UNIST), Ulsan 44919, Republic of Korea.}%
  \thanks{$^\dagger$Corresponding author:
    \href{mailto:sungil.kim@unist.ac.kr}{sungil.kim@unist.ac.kr}}%
}

\begin{document}

\maketitle
\thispagestyle{empty}
\pagestyle{empty}

\input{Abstract}

\input{Introduction}

\input{RelatedWorks.tex}

\input{Method.tex}

\input{Experiment.tex}

\input{Conclusion}

\addtolength{\textheight}{-12cm}   


\bibliographystyle{IEEEtran}  
\bibliography{ref}

\end{document}

%% file: Abstract.tex
\begin{abstract}
Contextual anomaly detection aims to identify abnormal behavior conditional on context variables, but practical deployments often face highly imbalanced context distributions where rare regimes can be critical information. Under such \emph{frequency bias}, context-conditioned models can produce unstable decisions and excessive false alarms in rare contexts. We propose \emph{Rarity-Gated Feature-wise Linear Modulation (RGFiLM)}, a rarity-aware conditioning module that combines feature-wise modulation (i.e., context-conditioned scaling and shifting of hidden features) with a gate controlled by a data-driven rarity score. The rarity score is estimated from the empirical distribution of context variables and regulates how strongly context modulates intermediate representations: the gate becomes more decisive under rare contexts while remaining conservative under frequent contexts. We evaluate RGFiLM on maritime trajectory anomaly detection using AIS motion sequences with ERA5 environmental context in an environment-sensitive detour scenario. When instantiated in a sequential anomaly scoring pipeline, RGFiLM achieves the best mean F1--False Positive Rate (FPR) trade-off among the compared context-agnostic and context-conditioned methods. These results suggest that explicitly accounting for context rarity is an effective approach for reducing false alarms in context-sensitive anomaly detection.
\end{abstract}

%% file: Introduction.tex
\section{Introduction}
\label{sec:intro}

Many anomalies are inherently \emph{contextual}: whether an observation is abnormal depends not only on its value but also on the context in which it occurs. A large trajectory deviation may be justified under severe weather yet anomalous in calm conditions. This motivates defining anomalies relative to a conditional distribution $p(\mathbf{b}|\mathbf{c})$, where $\mathbf{c}$ denotes context and $\mathbf{b}$ denotes behavior.

Contextual anomaly detection (CAD) applies this idea by separating context and behavior attributes and assessing whether behavior is typical given context. Existing CAD methods model $p(\mathbf{b}|\mathbf{c})$ using neighborhood-based estimates, conditional density models, conditional reconstruction, or quantile-based rules. However, a common limitation is that contexts are treated symmetrically once the model is trained, regardless of how frequently each context occurs.

In practice, context distributions are highly imbalanced: routine contexts are abundant, while rare contexts correspond to extreme or unusual regimes that are often safety-critical (e.g., severe weather, rare operating conditions)~\cite{kim2016ordinal}. We refer to the resulting failure mode as \emph{frequency bias}: models dominated by frequent contexts can mishandle behaviors under rare contexts, producing false alarms or unstable decisions. Prior work partially mitigates sparse contexts by backing off to a context-agnostic baseline when local contextual evidence is unreliable~\cite{ROCOD}, but this effectively down-weights rare contexts---precisely the regimes where contextual influence should be carefully controlled.

We address this gap with \emph{rarity-aware contextual conditioning}. We estimate a rarity score from the empirical context distribution and use it to control how strongly context modulates the representation used for anomaly scoring. Concretely, we propose a \emph{rarity-gated} conditioning module that interpolates between a context-agnostic representation and a context-modulated representation, becoming more decisive under rare contexts and more conservative under frequent contexts. In our implementation, the context-modulated representation is obtained via Feature-wise Linear Modulation (FiLM), which applies context-dependent feature-wise scaling and shifting to latent features.

In time-series anomaly detection, ``context'' is sometimes used to mean \emph{endogenous temporal structure} (e.g., seasonality or regime). In this paper, we use \emph{context} to denote exogenous conditioning variables (environmental/operational signals such as the fifth generation European Centre for Medium-Range Weather Forecasts (ECMWF) atmospheric reanalysis of the global climate (ERA5)), and refer to seasonality/regime effects as \emph{temporal context}.

We further consider a \emph{sequential, decision-based} anomaly setting, where anomalies arise as deviations in decision sequences (e.g., route changes or unusual control actions). We represent trajectories as $\tau=\{(s_t, e_t, a_t)\}_{t=1}^T$, where $s_t$ is state, $e_t$ is exogenous environment (context), and $a_t$ is action. A decision-based detector learns a policy $\pi(a_t| s_t, e_t)$ from normal trajectories and treats unlikely decisions as anomaly evidence. Our rarity-gated conditioning is modular and can be attached to such sequential decision models to regulate environmental influence under rare regimes.

As a case study, we apply the method to maritime trajectory anomaly detection using AIS motion sequences with ERA5 environmental context. We show that rarity-aware conditioning improves F1 while maintaining competitive precision and reducing false positive rate (FPR).

Our contributions are:
\begin{itemize}
    \item We identify \emph{frequency bias} under imbalanced context distributions and propose a \emph{rarity-aware conditioning} mechanism for contextual anomaly detection.
    \item We instantiate this idea as \emph{Rarity-Gated FiLM (RGFiLM)}, which interpolates between context-agnostic and FiLM-modulated representations with rarity-dependent decisiveness.
    \item In a maritime AIS--ERA5 case study, we show that rarity-aware conditioning improves the F1--FPR trade-off compared to standard conditioning baselines.
\end{itemize}

%% file: RelatedWorks.tex
\section{Related Work}
\label{sec:related}
\noindent\textbf{Maritime AIS anomaly detection.}
AIS-based maritime anomaly detection has leveraged real-time monitoring and statistical detection pipelines, including VAE--CUSUM monitoring~\cite{CUSUM}, grid-based Bayesian bootstrap detection~\cite{GRID}, and comparative evaluation of VAE-based monitoring statistics~\cite{VAE}. These methods provide practical surveillance tools but primarily score motion patterns via monitoring statistics and do not explicitly model \emph{context-conditioned} normality under imbalanced environmental regimes. This motivates our focus on contextual modeling and sequential decision-based detectors with rarity-calibrated context conditioning.

\subsection{Contextual Anomaly Detection and Rarity-Robust Context Conditioning}
Contextual anomaly detection (CAD) defines anomalies relative to the conditional distribution $p(\mathbf{b}|\mathbf{c})$ by separating context $\mathbf{c}$ and behavior $\mathbf{b}$. Early CAD formulations were developed primarily for tabular records and assume a user-specified context/behavior split, then model dependencies using parametric mappings or mixture models~\cite{CAD}. Later methods estimate $p(\mathbf{b}|\mathbf{c})$ using nonparametric neighborhood models, deep context-conditioned reconstruction, context-group discovery, or conditional interval/quantile rules (e.g., QCAD and related approaches)~\cite{ConOut,CADwithDirichlet,QCAD,CADwithKernel,DeepContextAD}. Although these methods do not directly capture temporal dependence, they establish the principle that abnormality should be judged \emph{conditional on context variables} rather than globally.

A key practical issue in CAD is that context distributions are often imbalanced: frequent contexts are well represented, while rare contexts are not. Under this imbalance, models can exhibit \emph{frequency bias}, producing unstable decisions or false alarms under rare contexts. ROCOD~\cite{ROCOD} explicitly discusses sparse contexts by combining a global behavior model with a neighbor-defined local model and backing off toward the global model when neighborhoods are sparse. While this improves stability, it effectively down-weights rare contexts. In contrast, we use rarity to \emph{calibrate} context conditioning rather than down-weight rare contexts.

Recent research adapts the CAD principle to multivariate time series by separating \emph{context variables} from target time-series variables and detecting anomalies relative to context-conditioned behavior over time~\cite{CADforMTS}. However, most approaches apply context conditioning uniformly and do not account for how rare a context is under the training distribution.
More broadly, in deep learning, explicit context variables are typically integrated into models via simple concatenation or structured modulation mechanisms such as Feature-wise Linear Modulation (FiLM), which generates feature-wise scaling and shifting from context to modulate intermediate representations~\cite{film_ori,FiLM}. These integration mechanisms implicitly assume sufficient coverage of the context distribution and can exhibit frequency bias under rare regimes~\cite{oh2025dualdynamics}. Our RGFiLM addresses this gap by making the strength of modulation explicitly dependent on a rarity score estimated from context variables, improving robustness and reducing false alarms in rare but safety-critical contexts.

\subsection{Sequential Decision-Based Anomaly Detection}
Many anomalies arise as deviations in \emph{decision sequences} rather than isolated observations (e.g., route changes or unusual control actions). Decision-based anomaly detection learns a policy that captures normal decision behavior and treats low-likelihood or inconsistent decisions as anomaly evidence~\cite{localOOD}. Reinforcement learning (RL) based approaches learn policies via reward maximization and detect deviations from the learned policy~\cite{RL}, while inverse reinforcement learning (IRL) based approaches infer rewards from demonstrations and use the induced policy for detection~\cite{IRL}. Although expressive, RL/IRL often require reward specification and either online interaction or a high-fidelity simulator, which is often impractical in safety-critical, purely observational settings.

Offline decision modeling (e.g., imitation learning) provides a practical alternative by learning from logged state--action trajectories without reward specification. In this work, we introduce a \emph{rarity-aware conditioning module} for sequential decision-based anomaly detection that modulates the latent representation used for decision modeling as a function of context variables and their rarity. We instantiate the anomaly scoring pipeline using OIL-AD~\cite{OILAD}, a representative offline imitation-learning based detector that derives anomaly evidence from deviations relative to a learned decision model, but the proposed rarity-gated conditioning is modular: it operates on intermediate representations and can be combined with diverse anomaly scoring heads (e.g., reconstruction-, forecasting-, density-, or decision-likelihood--based scoring).

%% file: Method.tex
\section{Method}
\label{sec:method}
\subsection{Overview}
Our goal is to perform anomaly detection under \emph{context variables} (e.g., environment/operating conditions) when the context distribution is imbalanced and rare regimes are safety-critical. We propose \emph{Rarity-Gated FiLM (RGFiLM)}, a modular rarity-aware conditioning module that operates on intermediate representations. RGFiLM can be attached to a wide range of sequential models and anomaly scoring heads; in our experiments, we instantiate it within a representative sequential anomaly detector trained on normal trajectories.

At each time step $t$, we observe a state $s_t\in\mathcal{S}$ (e.g., AIS-derived features), context variables $e_t\in\mathbb{R}^{d_e}$ (e.g., ERA5 environment), and (optionally) an action $a_t\in\mathcal{A}$, forming trajectories $\tau=\{(s_t,e_t,a_t)\}_{t=1}^T$. A sequence encoder produces a latent state representation
$$\mathbf{h}_t = f_{\text{state}}(s_{1:t}),$$
where $f_{\text{state}}$ can be any sequence backbone producing $\mathbf{h}_t$; RGFiLM is agnostic to this choice.

Context variables are injected through FiLM-style modulation. An environment encoder maps $e_t$ to feature-wise scaling and shifting parameters
$$(\boldsymbol{\gamma}_t, \boldsymbol{\beta}_t)=f_{\text{env}}(e_t),$$
yielding the context-modulated representation
$$\tilde{\mathbf{h}}_t=\boldsymbol{\gamma}_t\odot \mathbf{h}_t + \boldsymbol{\beta}_t,$$
where $\odot$ denotes element-wise multiplication.

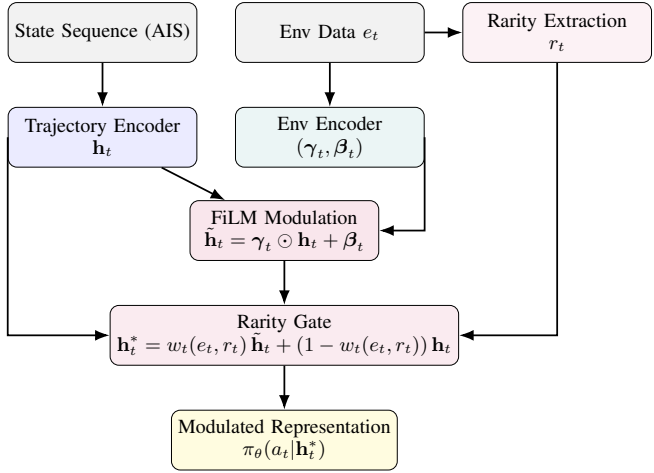
\begin{figure}[thbp]
    \centering
    \footnotesize
    \resizebox{0.48\textwidth}{!}{%
    \begin{tikzpicture}[
      node distance=7mm,
      every node/.style={font=\small},
      box/.style={draw, rounded corners, align=center, inner sep=3pt,
                  minimum width=3cm, minimum height=0.95cm},
      arr/.style={-Latex, thick}
    ]
      \node[box, fill=gray!10] (ais)  {State Sequence (AIS)};
      \node[box, fill=gray!10, right=6mm of ais] (era5) {Env Data $e_t$};

      \node[box, fill=blue!8, below=of ais]  (senc) {Trajectory Encoder \\ $\mathbf{h}_t$};
      \node[box, fill=teal!8, below=of era5] (eenc) {Env Encoder \\ $(\boldsymbol{\gamma}_t, \boldsymbol{\beta}_t)$};

      \node[box, fill=purple!5, right=6mm of era5] (rextract) {Rarity Extraction \\ $r_t$};

      \node[box, fill=purple!10, below=10mm of $(senc)!0.8!(eenc)$] (film)
        {FiLM Modulation \\
        $\displaystyle \tilde{\mathbf{h}}_t=\boldsymbol{\gamma}_t\odot\mathbf{h}_t+\boldsymbol{\beta}_t$};

      \node[box, fill=purple!8, below=of film] (rgate)
        {Rarity Gate \\
        $\displaystyle \mathbf{h}_t^* = w_t(e_t, r_t)\,\tilde{\mathbf{h}}_t + (1-w_t(e_t, r_t))\,\mathbf{h}_t$};

      \node[box, fill=yellow!15, below=of rgate] (hout) {Modulated Representation \\ $\pi_\theta(a_t | \mathbf{h}_t^*)$};

      \draw[arr] (ais)  -- (senc);
      \draw[arr] (era5) -- (eenc);
      \draw[arr] (era5) -- (rextract);
      \draw[arr] (senc) -- (film);
      \draw[arr] (eenc.east) |- (film);
      \draw[arr] (film) -- (rgate);
      \draw[arr] (senc.west) |- (rgate);
      \draw[arr] (rextract) |- (rgate);
      \draw[arr] (rgate) -- (hout);
    \end{tikzpicture}%
    }
    \caption{Overview of the proposed RGFiLM module}
    \label{fig:rarity-gated-film}
\end{figure}

Standard FiLM would pass $\tilde{\mathbf{h}}_t$ downstream directly. RGFiLM instead computes a rarity score $r_t\in[0,1]$ from the empirical context distribution and uses it to gate between $\mathbf{h}_t$ and $\tilde{\mathbf{h}}_t$, producing the final representation
$$\mathbf{h}_t^* = w_t(e_t,r_t)\,\tilde{\mathbf{h}}_t + \bigl(1-w_t(e_t,r_t)\bigr)\,\mathbf{h}_t,$$
which is then used by the downstream model (e.g., for reconstruction, forecasting, density, or decision-likelihood anomaly scoring). Figure~\ref{fig:rarity-gated-film} summarizes the overall architecture. 
In our experiments, we compute $r_t$ from significant wave height and wind speed, and exclude wind direction from the rarity computation.

\subsection{Rarity Extraction}
\label{sec:RarityExtraction}
The environmental vector $e_t \in \mathbb{R}^{d_e}$ (e.g., wave height, wind speed, wind direction) is used to quantify how rare the current environment is. First, we select a subset of environmental variables $e_t^r\in \mathbb{R}^{d_r}$. In our experiment setting, $e_t^r=[\text{wave height, wind speed}]$. We estimate the mean $\boldsymbol{\mu}$ and covariance $\boldsymbol{\Sigma}$ of $e_t^r$ from normal training samples and compute the Mahalanobis distance
\begin{equation}
    D_t^2 = (e_t^r - \boldsymbol{\mu})^{\top} \boldsymbol{\Sigma}^{-1} (e_t^r - \boldsymbol{\mu}).
    \label{eq:mahal}
\end{equation}
Assuming a multivariate Gaussian approximation, $D_t^2$ approximately follows a chi-squared distribution with $d_r$ degrees of freedom. We define the rarity score
\begin{equation}
    r_t = F_{\chi^2_{d_r}}(D_t^2),
    \label{eq:rarity}
\end{equation}
where $F_{\chi^2_{d_r}}$ is the corresponding CDF. Thus $r_t \in [0,1]$, with small values for common environments near $\boldsymbol{\mu}$ and large values for rare, extreme environments.

Other density-based scores could be used in place of \eqref{eq:mahal}--\eqref{eq:rarity}; we use the Mahalanobis-based rarity score due to its simplicity and efficiency.

\subsection{Rarity Gate}
The Rarity Gate controls how strongly the FiLM-modulated representation $\tilde{\mathbf{h}}_t$ influences the final state representation. It takes the environmental data $e_t$ and the rarity score $r_t$ as input.

First, we compute an intermediate score
$$s_t = g([e_t, r_t]),$$
where $g$ is a small neural network and $[\cdot,\cdot]$ denotes concatenation. We then apply a rarity-dependent temperature scaling:
$$\tilde{s}_t = \frac{s_t(1 + \kappa r_t)}{\tau},$$
with base temperature $\tau > 0$ and rarity sensitivity $\kappa \ge 0$. Finally, the gating weight and rarity-aware state representation are
\begin{equation}
    w_t = \sigma(\tilde{s}_t), \qquad
    \mathbf{h}_t^* = w_t\,\tilde{\mathbf{h}}_t + (1-w_t)\,\mathbf{h}_t,
    \label{eq:hstar}
\end{equation}
where $\sigma$ is the sigmoid function.

For a fixed value of $s_t$, the factor $(1 + \kappa r_t)/\tau$ grows with $r_t$ and thus amplifies the argument of the sigmoid, pushing $w_t$ closer to $0$ or $1$ as the environment becomes rarer. In other words, the gate becomes \emph{sharper} in rare environments, allowing the model to express a stronger preference for either the FiLM-modulated representation $\tilde{\mathbf{h}}_t$ or the original representation $\mathbf{h}_t$. In frequent environments, the effective temperature remains higher, so $w_t$ tends to vary more smoothly, producing softer mixtures and making it harder for the model to rely solely on extreme context effects in common regimes.

We refer to the overall mapping
$$\mathbf{h}_t^* = \mathrm{RGFiLM}(\mathbf{h}_t, e_t, r_t)$$
implemented by \eqref{eq:hstar} as the RGFiLM module.

\subsection{Integration with OIL-AD}
We integrate RGFiLM into the OIL-AD framework by replacing the original state representation with $\mathbf{h}_t^*$ wherever the state encoder output is used. Given normal trajectories $\{(s_t, e_t, a_t)\}$, the state encoder $f_{\text{state}}$, environment encoder $f_{\text{env}}$, gating network $g$, and the action and monotonicity heads of OIL-AD are trained jointly using the \emph{same} objective as in the original OIL-AD framework. That is, all loss components used in OIL-AD (the action loss and the monotonicity loss) are kept unchanged, except that the input state representation $\mathbf{h}_t$ is replaced by the rarity-aware, context-modulated representation $\mathbf{h}_t^*$.

Formally, if we denote the original OIL-AD objective by $\mathcal{L}_{\text{OIL-AD}}(\mathbf{h}_t)$, our model is trained by minimizing
$$\mathcal{L}_{\text{RGFiLM}} = \mathcal{L}_{\text{OIL-AD}}(\mathbf{h}_t^*),$$
with $\mathbf{h}_t^*$ given by the RGFiLM module. At test time, we also follow OIL-AD and compute anomaly scores from the same action and monotonicity based behavior features, with the key difference that these are now computed from $\mathbf{h}_t^*$ instead of the original state representation $\mathbf{h}_t$.

%% file: Experiment.tex
\section{Experiments}
\label{sec:experiments}

\subsection{Experimental Setup}

We evaluate RGFiLM in a setting where anomalies are inherently contextual and sequential. Our primary goal is to investigate whether rarity-aware conditioning can \emph{reduce false alarms} compared to both environment-agnostic and context-conditioned baselines, while maintaining or improving overall detection performance.  
Concretely, we ask:
\begin{itemize}
    \item In context-sensitive settings with imbalanced context variables (where rare contexts are safety-critical), does RGFiLM reduce false positives on normal data and improve the F1--FPR trade-off compared to context-agnostic and context-conditioned baselines?
\end{itemize}

To address this question, we construct an AIS-ERA5 dataset for a maritime case study and define a detour-based anomaly scenario in which route deviations are inherently context-dependent. Normal AIS trajectories already reflect environment-driven adjustments under harsh conditions; we therefore inject an additional lateral detour on top of these trajectories, yielding behaviors unlikely given the observed environmental context.
We evaluate detection performance using precision, recall, and F1 score on the detour scenario, and we explicitly quantify false alarms using the false positive rate (FPR) computed on normal trajectories.

\subsection{Dataset Construction}
\label{sec:dataset}
We construct an AIS--ERA5 dataset for the Bass Strait region (Australia) from July 2020 to December 2020, designed to support controlled lateral detour injection under diverse environmental conditions.

\subsubsection{AIS Data, and State--Action Representation}
AIS messages are sorted by timestamp for each vessel, cleaned, and temporally regularized following the OIL-AD preprocessing protocol.

\paragraph{Trajectory segmentation and filtering}
We split trajectories at long stationary segments using a destination-based heuristic: if a vessel remains below a speed threshold (SPEED $\le 1.2$) for more than a fixed number of consecutive steps, we treat this as a stop and split at the beginning and end of that stationary period. We retain voyage-like segments with length $T\in[50,300]$ and discard segments dominated by near-stationary behavior (e.g., $>80\%$ of points with very low speed).

\paragraph{State and action}
For each segment, we define the state as $s_t=[LAT_t,LON_t,SPEED_t]\in\mathbb{R}^3$. Actions are obtained by discretizing successive displacements into five motion directions (up, right, down, left, stay), yielding an aligned action sequence $\{a_t\}$ for imitation learning.
We split normal trajectories into train, validation, and test partitions at the trajectory level using a 70:10:20 ratio. Models are trained only on normal training trajectories, while injected anomalies are added only to the validation and test partitions for threshold selection and final evaluation.

\subsubsection{ERA5 Environmental Context and Rarity}
As exogenous context, we align each AIS point with ERA5 variables from the nearest time and grid cell: significant wave height, wind speed, and wind direction. Models are conditioned on the full context vector $e_t=[\mathrm{SWH}_t,\mathrm{WindSpeed}_t,\mathrm{WindDir}_t]$, while rarity is computed only from $e_t^r=[\mathrm{SWH}_t,\mathrm{WindSpeed}_t]$.

\subsection{Detour Scenario: Environment-Sensitive Anomalies}
\label{sec:detour-scenario}
We focus on an environment-sensitive anomaly setting where the \emph{same behavioral deviation} must be judged under \emph{different} environmental contexts. In the \textbf{detour scenario}, we model intentional abnormal route changes by synthetically modifying otherwise normal trajectories.

Trajectories are divided into train, validation and test sets at the trajectory level, and the model is trained only on unmodified (normal) trajectories. For the anomalous trajectories in the validation and test sets, we inject a detour by modifying a contiguous subsegment (≈2° eastward longitude shift) before rejoining the original path, while keeping the environmental sequence unchanged. Importantly, this injection is designed to be \textbf{environment-invariant}: the detour perturbation itself does not depend on wind/wave or any environmental variable. Consequently, the injected trajectory contains an \emph{abnormal detour behavior that is not explained by the environment}.

This design creates the central challenge we target. In real navigation, route changes can arise as \emph{environment-induced} responses (e.g., weather avoidance), so detour-like behaviors can be context-dependent and therefore ambiguous—especially under rare or extreme conditions where normal behavior is underrepresented. By injecting an \textbf{environment-invariant} detour while keeping the environmental sequence unchanged, we enforce a controlled setting where the \textbf{ground-truth anomaly mechanism is fixed} (the detour is not explained by the observed environment), yet the \textbf{difficulty of discrimination depends on context}. We evaluate overall detection performance on the full test set, and later analyze results by environmental rarity (Section~\ref{sec:detour-results}) to test robustness under rare contexts.

\begin{table*}[t]
    \centering
    \caption{Performance on the detour scenario over 20 runs (mean$\pm$std). Best scores in bold, second best underlined.}
    \label{tab:detour-results}
    \begin{tabular}{lcccc}
        \hline
        Model & F1 & Precision & Recall & False Positive Rate\\
        \hline
        OIL-AD~\cite{OILAD} & 0.524$\pm$0.060 & 0.401$\pm$0.056 & 0.782$\pm$0.143 & 0.120$\pm$0.032\\
        OIL-AD + Concat & 0.520$\pm$0.105 & 0.402$\pm$0.079 & 0.755$\pm$0.193 & 0.112$\pm$0.029\\
        OIL-AD + FiLM & {0.577$\pm$0.058} & \textbf{0.477$\pm$0.116} & {0.798$\pm$0.147} & \underline{0.098$\pm$0.044}\\
        OIL-AD + Gated FiLM & \underline{0.582$\pm$0.080} & {0.450$\pm$0.081} & \textbf{0.845$\pm$0.120} & {0.108$\pm$0.035}\\
        \hline
        ROCOD~\cite{ROCOD} & 0.172$\pm$0.054 & 0.116$\pm$0.040 & 0.475$\pm$0.299 & 0.365$\pm$0.238\\
        CAD for MTS~\cite{CADforMTS} & 0.346$\pm$0.074 & 0.335$\pm$0.132 & 0.461$\pm$0.174 & 0.120$\pm$0.074\\
        \hline
        RGFiLM (Ours) & \textbf{0.595$\pm$0.067} & \underline{0.474$\pm$0.080} & \underline{0.826$\pm$0.118} & \textbf{0.097$\pm$0.037}\\
        \hline
    \end{tabular}
\end{table*}

\subsection{Baselines and Training Setup}
\label{sec:baselines}

We compare the proposed Rarity-Gated FiLM against (i) OIL-AD-based variants that differ only in how context variables are incorporated, and (ii) two contextual anomaly detection baselines which are not sequential decision models.

\begin{itemize}
    \item \textbf{OIL-AD}~\cite{OILAD}:
    No environmental context; the sequential model operates solely on the latent state $\mathbf{h}_t$.

    \item \textbf{OIL-AD + Concat}:
    Context variables are concatenated with the state at each time step; the encoder consumes $[s_t, e_t]$. No FiLM or rarity gating is used.

    \item \textbf{OIL-AD + FiLM}:
    Standard FiLM conditioning~\cite{film_ori, FiLM}. The environment encoder produces $(\boldsymbol{\gamma}_t,\boldsymbol{\beta}_t)$ and the downstream heads consume $\tilde{\mathbf{h}}_t$ without rarity gating.

    \item \textbf{OIL-AD + Gated FiLM (no rarity)}:
    FiLM modulation is combined with a learnable gate between $\mathbf{h}_t$ and $\tilde{\mathbf{h}}_t$, but the gate \emph{does not use} the rarity score (equivalently, $r_t$ is fixed to $0$). This ablation tests the effect of gating alone without rarity-awareness.
    
    \item \textbf{ROCOD}~\cite{ROCOD}:
    A contextual outlier detection baseline that combines local (context-neighbor) and global (context-agnostic) behavior models. 

    \item \textbf{CAD for MTS}~\cite{CADforMTS}:
    A contextual anomaly detection method for multivariate time series that uses explicit context variables. 
    
    \item \textbf{Rarity-Gated FiLM (Ours)}:
    The proposed RGFiLM module produces $\mathbf{h}_t^*$ via rarity-gated interpolation between $\mathbf{h}_t$ and $\tilde{\mathbf{h}}_t$ (Section~\ref{sec:method}).
\end{itemize}

For OIL-AD-based variants, we use the same encoder architecture and the same optimization settings (batch size, learning rate, epochs) to isolate the effect of the conditioning mechanism. For RGFiLM, we set the base temperature to $\tau=1.2$ and rarity sensitivity to $\kappa=0.1$, selected on the validation set. All anomaly scores for OIL-AD-based variants are computed using the same scoring pipeline, with $\mathbf{h}_t$ replaced by $\mathbf{h}_t^*$ where applicable.

\subsection{Results on the Detour Scenario}
\label{sec:detour-results}
Table~\ref{tab:detour-results} reports false positive rate, precision, recall, and F1 score for all methods on the detour scenario.

We repeat the experiment over 20 random seeds. In each run, the validation and test sets are constructed with a 10:1 normal-to-anomaly ratio, where each trajectory is a variable-length sequence $\tau=\{(s_t,e_t,a_t)\}_{t=1}^T$. The detector produces time-step anomaly evidence from sequential decision signals, and we form a trajectory-level anomaly score by mean aggregation, $S(\tau)=\frac{1}{T}\sum_{t=1}^{T} z_t$. The anomaly threshold is selected on the validation set and used for test evaluation. We report the mean$\pm$std over all runs.

Table~\ref{tab:detour-results} shows that RGFiLM achieves the best mean F1--FPR trade-off, obtaining the highest mean F1 score (0.595) and the lowest mean false positive rate (0.097). Compared with the context-agnostic OIL-AD baseline, RGFiLM improves F1 (0.524$\rightarrow$0.595) while reducing FPR (0.120$\rightarrow$0.097). Compared with OIL-AD + FiLM, RGFiLM achieves a higher mean F1 (0.577$\rightarrow$0.595) and recall (0.798$\rightarrow$0.826), with comparable FPR (0.098$\rightarrow$0.097). Compared with OIL-AD + Gated FiLM, RGFiLM improves mean F1 (0.582$\rightarrow$0.595) and reduces FPR (0.108$\rightarrow$0.097), although Gated FiLM attains the highest recall. These results indicate that rarity-aware gating improves the F1--FPR balance, with its most reliable gains appearing in false-alarm reduction.

We further perform paired Wilcoxon signed-rank tests over the 20 matched seeds. RGFiLM significantly improves both F1 and FPR over OIL-AD and Concat ($p<0.05$ for all), and significantly reduces FPR over OIL-AD + Gated FiLM ($p=0.028$). The significant gains are concentrated on false-alarm reduction against baselines with higher FPR, while FiLM already attains a similar mean FPR to RGFiLM (0.098 vs.\ 0.097). 

Among OIL-AD variants, {OIL-AD + Concat} does not improve over the context-agnostic baseline in mean F1, suggesting that naive input-level fusion is insufficient for effectively using environmental variables. {OIL-AD + FiLM} and {OIL-AD + Gated FiLM} both improve over {OIL-AD + Concat}, indicating that feature-wise environmental conditioning and adaptive gating are useful. Building on these, the rarity-aware gate provides its clearest additional benefit on false alarms: relative to OIL-AD + Gated FiLM, which uses the same gating mechanism without rarity, RGFiLM significantly reduces FPR while attaining higher mean F1. Because this comparison differs only in rarity-awareness, it isolates the contribution of the rarity gate beyond gating alone. External reference baselines (ROCOD and CAD for MTS) show substantially lower F1, indicating that in this sequential decision-based trajectory setting the sequential decision-model family is better aligned with the task than non-policy contextual baselines.

\begin{figure*}[htbp]
    \centering
    \includegraphics[width=0.78\textwidth]{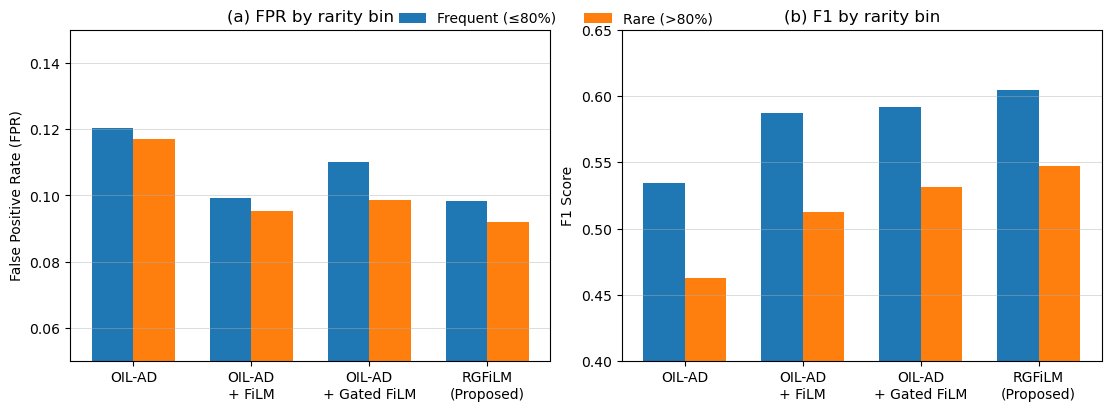}
    \caption{Ablation by environmental rarity. (a) FPR and (b) F1 score on frequent and rare trajectories, where rare trajectories are the 20\% most environmentally rare  according to the trajectory-level rarity score}
    \label{fig:rarity-bin-ablation}
\end{figure*}

\noindent\textbf{Ablation study across rarity bins.}
Figure~\ref{fig:rarity-bin-ablation} reports FPR and F1 stratified by environmental rarity (frequent: bottom 80\%, rare: top 20\%). 

All methods show lower F1 under rare contexts than under frequent contexts. RGFiLM attains the highest mean F1 in both the frequent and rare regimes (0.604 and 0.547) and the lowest mean FPR in both (0.098 and 0.092). Its mean F1 advantage over FiLM is larger in the rare regime (0.513$\rightarrow$0.547) than in the frequent regime (0.587$\rightarrow$0.604). This descriptive trend suggests that the benefit of rarity-aware gating is more visible in mean performance when contexts are underrepresented. RGFiLM therefore retains the strongest mean performance across both regimes while showing its largest mean performance gains in rare contexts.

%% file: Conclusion.tex
\section{Conclusion}
\label{sec:conclusion}
This paper revisited contextual anomaly detection from a \emph{rarity-aware} perspective and  proposed \emph{Rarity-Gated FiLM (RGFiLM)}, a modular conditioning mechanism that combines FiLM-style context modulation with a rarity-dependent gate. By estimating a rarity score from the empirical distribution of context variables, RGFiLM controls how strongly context affects intermediate representations: the gate becomes more decisive under rare contexts and remains conservative under frequent contexts, mitigating frequency bias and reducing false alarms. 

We evaluated RGFiLM on maritime trajectory anomaly detection using an AIS--ERA5 dataset and a detour-based anomaly scenario. When instantiated in a sequential anomaly scoring pipeline, RGFiLM achieved the best mean F1--FPR trade-off among the context-agnostic and context-conditioned methods. These results indicate that explicitly accounting for context rarity can improve robustness in context-sensitive anomaly detection. 

Our evaluation uses synthetic detour injection with fixed magnitude, which may not cover the full diversity of real-world anomaly mechanisms. Rarity is estimated from the empirical context distribution and may be sensitive to limited coverage or domain shift in environmental variables. Future work will extend the approach to broader anomaly types, regions, and operational periods.

\section*{Acknowledgment}
This work was supported by the National Research Foundation of Korea (NRF) grant funded by the Korea government(MSIT)(No.RS-2023-00218913).

%% file: ref.bib
@article{CAD,
  title={Conditional anomaly detection},
  author={Song, Xiuyao and Wu, Mingxi and Jermaine, Christopher and Ranka, Sanjay},
  journal={IEEE Transactions on knowledge and Data Engineering},
  volume={19},
  number={5},
  pages={631--645},
  year={2007},
  publisher={IEEE}
}

@article{kim2016ordinal,
  title={Ordinal classification of imbalanced data with application in emergency and disaster information services},
  author={Kim, Sungil and Kim, Heeyoung and Namkoong, Younghwan},
  journal={IEEE Intelligent Systems},
  volume={31},
  number={5},
  pages={50--56},
  year={2016},
  publisher={IEEE}
}

@inproceedings{oh2025dualdynamics,
  title={DualDynamics: Synergizing implicit and explicit methods for robust irregular time series analysis},
  author={Oh, YongKyung and Lim, Dong-Young and Kim, Sungil},
  booktitle={Proceedings of the AAAI Conference on Artificial Intelligence},
  volume={39},
  number={18},
  pages={19730--19739},
  year={2025}
}

@inproceedings{ROCOD,
  title={Robust contextual outlier detection: Where context meets sparsity},
  author={Liang, Jiongqian and Parthasarathy, Srinivasan},
  booktitle={Proceedings of the 25th ACM international on conference on information and knowledge management},
  pages={2167--2172},
  year={2016}
}

@article{CADforMTS,
  title={Contextual anomaly detection for multivariate time series data},
  author={Kim, Hyojoong and Kim, Heeyoung},
  journal={Quality Engineering},
  volume={35},
  number={4},
  pages={686--695},
  year={2023},
  publisher={Taylor \& Francis}
}

@inproceedings{ConOut,
  title={Conout: Contextual outlier detection with multiple contexts: Application to ad fraud},
  author={Meghanath, MY and Pai, Deepak and Akoglu, Leman},
  booktitle={Joint European Conference on Machine Learning and Knowledge Discovery in Databases},
  pages={139--156},
  year={2018},
  organization={Springer}
}

@article{QCAD,
  title={Explainable contextual anomaly detection using quantile regression forests: Z. Li, M. van Leeuwen},
  author={Li, Zhong and Van Leeuwen, Matthijs},
  journal={Data Mining and Knowledge Discovery},
  volume={37},
  number={6},
  pages={2517--2563},
  year={2023},
  publisher={Springer}
}

@article{DeepContextAD,
  title={Deep Context-Conditioned Anomaly Detection for Tabular Data},
  author={King, Spencer and Zhang, Zhilu and Yu, Ruofan and Coskun, Baris and Ding, Wei and Cui, Qian},
  journal={arXiv e-prints},
  pages={arXiv--2509},
  year={2025}
}

@article{OILAD,
  title={Oil-ad: An anomaly detection framework for decision-making sequences},
  author={Wang, Chen and Erfani, Sarah and Alpcan, Tansu and Leckie, Christopher},
  journal={Pattern Recognition},
  volume={166},
  pages={111656},
  year={2025},
  publisher={Elsevier}
}

@inproceedings{IRL,
  title={Sequential anomaly detection using inverse reinforcement learning},
  author={Oh, Min-hwan and Iyengar, Garud},
  booktitle={Proceedings of the 25th ACM SIGKDD International Conference on Knowledge Discovery \& data mining},
  pages={1480--1490},
  year={2019}
}

@inproceedings{RL,
  title={Rlad: Reinforcement learning based anomaly detection system for iot devices in smart homes},
  author={Chadha, Angela Raj and Reddy, Kethamreddy Karthikeya and Sountharrajan, S},
  booktitle={2024 15th International Conference on Computing Communication and Networking Technologies (ICCCNT)},
  pages={1--7},
  year={2024},
  organization={IEEE}
}

@article{FiLM,
  title={User-conditioned neural control policies for mobile robotics},
  author={Bauersfeld, Leonard and Kaufmann, Elia and Scaramuzza, Davide},
  journal={arXiv preprint arXiv:2211.12181},
  year={2022}
}

@inproceedings{film_ori,
  title={Film: Visual reasoning with a general conditioning layer},
  author={Perez, Ethan and Strub, Florian and De Vries, Harm and Dumoulin, Vincent and Courville, Aaron},
  booktitle={Proceedings of the AAAI conference on artificial intelligence},
  volume={32},
  number={1},
  year={2018}
}

@article{localOOD,
  title={Locally most powerful bayesian test for out-of-distribution detection using deep generative models},
  author={Kim, Keunseo and Shin, JunCheol and Kim, Heeyoung},
  journal={Advances in Neural Information Processing Systems},
  volume={34},
  pages={14913--14924},
  year={2021}
}

@article{CADwithDirichlet,
  title={Contextual anomaly detection for high-dimensional data using Dirichlet process variational autoencoder},
  author={Kim, Hyojoong and Kim, Heeyoung},
  journal={IISE Transactions},
  volume={55},
  number={5},
  pages={433--444},
  year={2023},
  publisher={Taylor \& Francis}
}

@article{CADwithKernel,
  title={Deep embedding kernel mixture networks for conditional anomaly detection in high-dimensional data},
  author={Kim, Hyojoong and Kim, Heeyoung},
  journal={International Journal of Production Research},
  volume={61},
  number={4},
  pages={1101--1113},
  year={2023},
  publisher={Taylor \& Francis}
}

@article{VAE,
  title={Comparative evaluation of VAE-based monitoring statistics for real-time anomaly detection in AIS data},
  author={Oh, YongKyung and Yoon, Kwonin and Park, Jaemin and Kim, Sungil},
  journal={Maritime Policy \& Management},
  volume={52},
  number={4},
  pages={609--626},
  year={2025},
  publisher={Taylor \& Francis}
}

@article{GRID,
  title={Grid-based Bayesian bootstrap approach for real-time detection of abnormal vessel behaviors from AIS data in maritime logistics},
  author={Oh, YongKyung and Kim, Sungil},
  journal={IEEE Transactions on Automation Science and Engineering},
  volume={21},
  number={4},
  pages={6680--6692},
  year={2023},
  publisher={IEEE}
}

@article{CUSUM,
  title={Maritime anomaly detection based on VAE-CUSUM monitoring system},
  author={Park, Jaemin and Kim, Sungil},
  journal={Journal of the Korean Institute of Industrial Engineers},
  volume={46},
  number={4},
  pages={432--442},
  year={2020}
}
